# Active Learning for Developing Personalized Treatment

**Kun Deng**
Department of Statistics
University of Michigan
kundeng@umich.edu

**Joelle Pineau**
Department of Computer Science
McGill University
jpineau@cs.mcgill.ca

**Susan Murphy**
Department of Statistics
University of Michigan
samurphy@umich.edu

## Abstract

The personalization of treatment via biomarkers and other risk categories has drawn increasing interest among clinical scientists. Personalized treatment strategies can be learned using data from clinical trials, but such trials are very costly to run. This paper explores the use of active learning techniques to design more efficient trials, addressing issues such as whom to recruit, at what point in the trial, and which treatment to assign, throughout the duration of the trial. We propose a minimax bandit model with two different optimization criteria, and discuss the computational challenges and issues pertaining to this approach. We evaluate our active learning policies using both simulated data, and data modeled after a clinical trial for treating depressed individuals, and contrast our methods with other plausible active learning policies.

## 1 Introduction

The heterogeneity of responses to treatment has become a highly influential factor in shaping future drug development and clinical practices [13]. Already, some biomarkers (e.g. findings in the brain MRI) and risk categories (e.g. gender or smoking history) are known to be strong indicators of differential responses to treatments. Clinical scientists are very interested in the development of novel methods to identify subpopulations (i.e. patients stratified based on some biomarkers) that respond differently to treatment, and correspondingly, the best treatments for those subpopulations.

Formally, the goal is to learn good Individualized Treatment Rules (ITR). ITRs are mappings from the pretreatment observations to treatments. A major challenge is how best to collect data necessary to learn good ITRs. The most common approach is to run a standard Randomized Clinical Trial (RCT), in which the number of patients in each subpopulation reflects the relative sizes of each subpopulation, followed by a stratified data analysis in the hopes of revealing the best treatment suited for each subpopulation. However, there are at least two disadvantages to this approach. First, it usually lacks the power to discern the treatment effect for many of the subpopulations, especially when there exist some subpopulations that are rare or the treatment effects for them are relatively small. Second, such an approach will also waste trial resources on subpopulations which experience treatment effects that are relatively large. This second disadvantage is particularly problematic, given the high cost of running clinical trials. In view of these problems, a more sensible design for developing personalized treatment would be to adapt the number of patients recruited from each subpopulation, as demanded by the goal of the trial.

The problem of trial design can be formulated as a multi-armed bandit [11], whereby each arm corresponds to a (subpopulation, treatment) pair. Each decision step corresponds to the recruitment and treatment of a new subject. In this paper, we propose a Minimax bandit model that intelligently recruits patients from different subpopulations and assigns them to different treatments, in order to optimize the quality of the constructed ITR. We formalize two different optimization criteria related to the quality of an ITR, and propose exploration policies for solving them. We present experimental results comparing the performance of our approach to alternative active learning strategies. We observe that our approach can successfully balance the objectives of the clinical trial, including recruiting subjects from subgroups for which there is a high potential of finding good ITRs. The techniques presented in the paper generalize to a large spectrum of other domains requiring efficient contextual exploration, including experimental design, drug discovery, and the design of personalized interfaces.

## 2 Methods and Algorithms

Consider the following motivating example, in which one must decide between two treatment options ($a_1$ and $a_2$) for treating subjects from four subpopulations ($c_1, \ldots, c_4$). We consider perhaps the simplest goal, that of assigning each subpopulation to the most effective of the two competing treatments, ignoring other characteristics/covariates of the subjects. This goal, though simple, has recently been of much interest [14], particularly in the area of stratified medicine [13]. We denote the mean response under each treatment for subpopulation $c_i$ to be $\mu_{i1}$ and $\mu_{i2}$. So a treatment assignment may look like this:

$$d(c_i) = \begin{cases} a_1 & \text{if } \hat{\mu}_{i1} - \hat{\mu}_{i2} \geq 0 \\ a_2 & \text{if } \hat{\mu}_{i1} - \hat{\mu}_{i2} < 0 \end{cases} \quad \forall i \in \{1, 2, 3, 4\} \tag{1}$$

where the $\hat{\mu}_{i\cdot}$ are the estimates of $\mu_{i\cdot}$. The estimator of the $i$-th subpopulation's treatment effect is $\hat{\mu}_{i1} - \hat{\mu}_{i2}$.

The above treatment rule can be regarded as a special case of a more general construct, called an Individualized Treatment Rule (ITR) [10]. Formally, for each patient, we have the pretreatment observation $X \in \mathcal{X}$, summarizing various aspects of individual heterogeneity, treatment $A$ taking values in a finite, discrete treatment space $\mathcal{A}$, and a real-valued response $R$ (assuming large values are desirable). An Individualized Treatment Rule, denoted $d$, is a deterministic decision function from $\mathcal{X}$ into the treatment space $\mathcal{A}$. We aim to construct this rule so as to maximize the future response $R$. In the above example, $X = \{1, 2, 3, 4\}$ is the set of subpopulations and $A = \{1, 2\}$ is the set of treatments.

Our goal is to develop an efficient active learning policy for learning reliable ITRs that inform clinical decision making. We assume that the constraints of the trial do not prevent recruiting patients from specific subpopulations. We also assume that the active treatment period for a patient is of relatively short duration compared to the pace of patient recruitment. Finally, we assume that once a patient is recruited into the trial, the treatment and monitoring process are costly, and thus there is a constrained budget (say $N$ subjects) for the entire trial.

Within this framework, there are many possible objectives that can be used to formalize the problem of online exploration for learning ITRs. For the purposes of this paper, we focus on two of these:

- Minimize the overall uncertainty about the treatment effects for all subpopulations.
- Minimize the overall error of incorrectly selecting a nonoptimal treatment for all subpopulations.

We discuss other possible objectives, and issues related to these choices in Section 5.

The model we use is formally described as follows: there are $C$ bandits (corresponding to the $C$ subpopulations), each equipped with $K$ arms (corresponding to the $K$ treatments). At each time step (corresponding to the recruitment of a patient), we are only allowed to select one bandit. For that bandit, we need to further choose an arm to pull. There will be a total of $N$ pulls. We assume that the response to pulling the $j$th arm of the $i$th bandit, denoted as $(i, j)$, follows the distribution $D_{ij}$ , with mean $\mu_{ij}$ (corresponding to the response to treatment $j$ in the $i$th subpopulation). At each time point $n$, the estimated mean response of arm $j$ for the $i$th bandit is $\hat{\mu}^n_{ij}$, and the loss for bandit $i$ is measured by a known loss function $L^n_i$. The overall loss of an online active learning policy $\pi$ is measured by the worst-case loss over the $C$ bandits:

$$L^n(\pi) = \max_{1 \leq i \leq C} L^n_i. \tag{2}$$

We would like to design $\pi$ such that the loss $L^n(\pi)$ is small. In this paper, we consider two concrete loss functions $L^n_i$. They are defined in the next two subsections.

### 2.1 Minimizing the maximal variance of the estimated treatment effects

In this section, we develop an active learning policy for a common case in which there are only two treatments per subpopulation ($K = 2$). Assuming the patients in the subpopulations respond independently, we have that the variance of the estimated treatment effect $\mathbb{V}[\hat{\mu}_{i1} - \hat{\mu}_{i2}] = \mathbb{V}[\hat{\mu}_{i1}] + \mathbb{V}[\hat{\mu}_{i2}]$. Thus we consider the case where $L^n_i = \sum_{j=\{1,2\}} \mathbb{V}[\hat{\mu}^n_{ij}]$, is the variance of the estimated treatment effect for subpopulation $i$. The corresponding $L^n(\pi)$ loss refers to the worst uncertainty about the ITR over the different subpopulations. In the clinical setting, optimizing this criterion allows us to distribute the resources wisely so that the learned ITR has a bounded uncertainty (or vice versa, high confidence) for each subpopulation. This is in contrast to the approach of using a standard randomized clinical trial, in which the number of patients in each subpopulation reflects the relative sizes of each subpopulation, and which use equal randomization of the treatments, thus running the risk of yielding highly variable estimated treatment effects for some subpopulations.

We base our active learning policy for this optimization criterion on an optimal "oracle" allocation policy, that has been endowed with the knowledge of the variances $\sigma^2_{ij}$ of the responses. Note that if an arm $(i, j)$ has been pulled $n_{ij}$ times, then $\mathbb{V}[\hat{\mu}_{ij}] = \frac{\sigma^2_{ij}}{n_{ij}}$. Re-call that there are a total of $N$ pulls. The optimal oracle allocation can be computed by solving the following convex optimization problem, ignoring integer constraints on $n_{ij}$:

$$\begin{aligned} \underset{n_{ij}}{\text{minimize}} \quad & \max_i \textstyle\sum_j \frac{\sigma_{ij}^2}{n_{ij}} \\ \text{s.t.} \quad & \textstyle\sum_i \sum_j n_{ij} = N \\ & n_{ij} > 0 \qquad \forall i \; \forall j \end{aligned} \tag{3}$$

The optimal allocation for the $i$th bandit and $j$th arm, ignoring integer constraints, is

$$n_{ij}^* = \frac{\sigma_{ij} \sum_j \sigma_{ij}}{\sum_i (\sum_j \sigma_{ij})^2} N \propto \frac{\sigma_{ij} \sum_j \sigma_{ij}}{\sum_i (\sum_j \sigma_{ij})^2} \tag{4}$$

To see this, note that the above convex optimization problem can be restated as:

$$\begin{aligned} \underset{n_{ij}, r}{\text{minimize}} \quad & r \\ \text{s.t.} \quad & \textstyle\sum_j \frac{\sigma_{ij}^2}{n_{ij}} \le r \qquad \forall i \in \{1, ..., C\} \\ & \textstyle\sum_i \sum_j n_{ij} = N \\ & -n_{ij} \le 0 \qquad \forall i \in \{1, ..., C\}, j \in \{1, ..., K\} \end{aligned}$$

The full Lagrangian is

$$\begin{aligned} L(n_{ij}, r, \lambda_{ij}, \alpha, \beta_{ij}) &= r + \sum_i \lambda_i (\sum_j \frac{\sigma_{ij}^2}{n_{ij}} - r) \\ &+ \alpha (\sum_i \sum_j n_{ij} - N) + \sum_i \sum_j \beta_{ij} (-n_{ij}) \end{aligned}$$

By the KKT condition, $\beta_{ij} = 0$ as $n_{ij}$ has to be strictly positive ; $\frac{\partial L}{\partial n_{ij}} = 0$ yields

$$\alpha = \frac{\lambda_i \sigma_{ij}^2}{n_{ij}^2}$$

for all $i$ and $j$; $\frac{\partial L}{\partial r} = 0$ yields

$$1 - \sum_i \lambda_i = 0,$$

and finally, $\alpha(\sum_i \sum_j n_{ij} - N) = 0$. It's easy to verify that $n_{ij}^*$ in (4) and $r^* = \frac{\sum_i (\sum_j \sigma_{ij})^2}{N}$ are solutions. Thus, the optimal oracle allocation would recruit $n_{ij}^*$ subjects from subpopulation $i$ and assign them treatment $j$.

Our proposed active learning policy, called Adaptive Randomization with Estimated Optimal Allocation (AREOA), uses (4) as the basis for selecting subjects and assigning treatments. It is worth noting that $\left\{ \frac{\sigma_{ij} \sum_j \sigma_{ij}}{\sum_i (\sum_j \sigma_{ij})^2}; i \in \{1, ..., C\}, j \in \{1, ..., K\} \right\}$ forms a proper probability distribution, so if an active learning policy samples according to this distribution at each of the $N$ decision points, it will end up allocating in expectation $n_{ij}^*$ for each arm $(i, j)$. Of course, the values of $\sigma_{ij}$ are unknown in an online setting, however they can be estimated via the sample standard deviation, $\hat{\sigma}_{ij}$, when there is sufficient data. Also, according to (4), when either $\sigma_{ij}$ or $\sum_j \sigma_{ij}$ is large, arm $(i, j)$ or bandit $j$ should be pulled more often, which is consistent with our objective. The details of this active learning policy (together with two other policies that will be used for comparison later) are described in Figure 1. Note that to make AREOA more robust, our implementation of AREOA starts with an initial exploratory phase that uses a fixed portion of budget to estimate the inital values for $\hat{\sigma}$, then uses the main optimality criterion (4) in the context of an epsilon-greedy strategy that maintains a small probability of uniform random exploration.

Choose each treatment for each subpopulation $B$ times in the first $B \times C \times K$ trials where $C$ and $K$ are number of subpopulations and number of treatments.
Set $n_{ij}^{(n)} = B$ and $n = BCK$
**while** $n \le N$ **do**
  Compute the standard error estimate $\hat{\sigma}_{ij}^{(n)}$ for treatment $(i, j)$ at time point $n$
  *Option 1:AREOA.*
  **if** $\sum_i \left( \sum_j \hat{\sigma}_{ij}^{(n)} \right)^2 \neq 0$ **then**
    Let $\tau_{ij} = \frac{\hat{\sigma}_{ij}^{(n)} \sum_j \hat{\sigma}_{ij}^{(n)}}{Z}$ where $Z$ is chosen such that $\sum_i \sum_j \tau_{ij} = 1$
  **else**
    let $\tau_{ij} = \frac{1}{CK}$
  **end if**
  Pick the next subpopulation and treatment pair $(i, j)$ with probability $(1 - \epsilon)\tau_{ij} + \epsilon * \frac{1}{CK}$
  *Option 2:AARandom.*
  Randomly pick a subpopulation $i$ according to its composition in the general population, and then pick a treatment $j$ uniformly at random.
  *Option 3:GAFS-MAX.*
  Assume some arbitrary but fixed ordering for the set of all $(i, j)$ pairs.
  Let $U_n = \{(k, l) : n_{k,l}^{(n)} < \sqrt{n} + 1\}$
  Let

$$I_{n+1} = \begin{cases} \min U_n & \text{if } U_n \neq \varnothing \\ \arg\max \frac{(\hat{\sigma}_{ij}^{(n)})^2}{n_{ij}^{(n)}} & \text{otherwise} \end{cases}$$

  Choose option $I_{n+1}$ and update $n_{ij}^{(n+1)}$ accordingly
  $n = n + 1$
**end while**

Figure 1: Algorithm Framework for minimizing the maximal variance. AREOA is the proposed active learning policy. AARandom and GAFS-MAX are two alternative active learning policies.

## 2.2 Minimizing the maximal probability of incorrectly selecting suboptimal treatments

Next, we consider a second objective, namely minimizing the probability of selecting an incorrect treatment.

Assume that there is a single best treatment for each subpopulation; let $j^*$ [1] be the index of the true best treatment. Define

$$L_i^n = \Pr[\max_{j \neq j^*} \hat{\mu}_{ij} \geq \hat{\mu}_{ij^*}] \ ,$$

the probability that an inferior treatment has equal or higher estimated mean response for subpopulation $i$ than that of the true best. The loss $L^n(\pi) = \max_{1 \leq i \leq C} L_i^n$ aims to control the maximal error of incorrectly selecting a suboptimal treatment based on the data. Again, we base our adaptive learning policy on an optimal oracle allocation that assumes known subpopulation means and variances:

$$\begin{array}{lr} \text{minimize} \quad \max_i \Pr[\max_{j \neq j^*} \hat{\mu}_{ij} \geq \hat{\mu}_{ij^*}] & (5) \\ \text{s.t.} \quad \sum_i \sum_j n_{ij} = N. & \end{array}$$

Using the standard notation $\Phi(\cdot)$ and $\phi(\cdot)$ for the cumulative density and density function of standard normal random variables, we have

$$\begin{aligned} L_i &\stackrel{def}{=} L(\mathbf{n}_{i,\cdot}; \boldsymbol{\mu}_{i,\cdot}, \boldsymbol{\sigma}_{i,\cdot}) \stackrel{def}{=} \Pr[\max_{j \neq j^*} \hat{\mu}_{ij} \geq \hat{\mu}_{ij^*}] \\ &= 1 - \Pr\left[\max_{j \neq j^*} \hat{\mu}_{ij} < \hat{\mu}_{ij^*}\right] \\ &= 1 - \mathbb{E}\left[\Pr\left[\max_j \hat{\mu}_{ij} < \hat{\mu}_{ij^*} | \hat{\mu}_{ij^*}\right]\right] \\ &= 1 - \mathbb{E}\left[\Pi_{j \neq j^*} \Pr\left[\hat{\mu_{ij}} < \hat{\mu}_{ij^*} | \hat{\mu}_{ij^*}\right]\right] \\ &= 1 - \mathbb{E}\left[\Pi_{j \neq j^*} \Phi(\frac{\hat{\mu}_{ij^*} - \mu_{ij}}{\sigma_{ij}/\sqrt{n_{ij}}})\right] \\ &= 1 - \int \Pi_{j \neq j^*} \Phi(\frac{x - \mu_{ij}}{\sigma_{ij}/\sqrt{n_{ij}}}) \phi(\frac{x - \mu_{ij^*}}{\sigma_{ij^*}/\sqrt{n_{ij^*}}}) \frac{1}{\sigma_{ij^*}/\sqrt{n_{ij^*}}} dx \\ &= 1 - \int \Pi_{j \neq j^*} \Phi(\frac{\frac{\sigma_{ij^*}}{\sqrt{n_{ij^*}}} z + \mu_{ij^*} - \mu_{ij}}{\sigma_{ij}/\sqrt{n_{ij}}}) \phi(z) dz, \qquad (6) \end{aligned}$$

which yields a closed form for $L_i$. However, $L_i$ is not convex in $\boldsymbol{n_{i\cdot}}$, neither is $\max_i L_i$.

We use the following relaxation to derive a surrogate for the original objective function:

$$\Pr[\max_{j \neq j^*} \hat{\mu}_{ij} \geq \hat{\mu}_{ij^*}] \leq \sum_{j \neq j^*} \Pr\left[\hat{\mu}_{ij} \geq \hat{\mu}_{ij^*}\right] \leq \sum_{j \neq j^*} \frac{\mathbb{V}(\hat{\mu}_{ij} - \hat{\mu}_{ij^*})}{(\mu_{ij} - \mu_{ij^*})^2},$$

where the first inequality is due to Boole's inequality, and the second inequality is due to Chebyshev's inequality. We solve the following convex optimization problem (ignoring integer constraints):

$$\begin{array}{lr} \text{surrogate:} \min_{n_{ij}}\text{imize} \quad \max_i \sum_{j \neq j^*} \frac{\frac{\sigma_{ij}^2}{n_{ij}} + \frac{\sigma_{ij^*}^2}{n_{ij^*}}}{(\mu_{ij} - \mu_{ij^*})^2} & (7) \\ \text{s.t.} \quad \sum n_{ij} = N. & \end{array}$$

[1] For brevity, we abused the notation slightly, in that $j^*$ depends on $i$, and should really be written as $j_i^*$.

Table 1: Datasets for the AREOA comparisons

| DS | subpop./ treatments | dist. | means | variances |
|---|---|---|---|---|
| DS1 | 4/2 | $\begin{pmatrix} .25 \\ .25 \\ .25 \\ .25 \end{pmatrix}$ | $\begin{pmatrix} 1 & 4 \\ 2 & 2 \\ 4 & 1 \\ 2 & 2 \end{pmatrix}$ | $\begin{pmatrix} 1000 & 1000 \\ 100 & 100 \\ 100 & 100 \\ 100 & 100 \end{pmatrix}$ |
| DS2 | 4/2 | $\begin{pmatrix} .1 \\ .3 \\ .3 \\ .3 \end{pmatrix}$ | $\begin{pmatrix} 1 & 4 \\ 2 & 2 \\ 4 & 1 \\ 2 & 2 \end{pmatrix}$ | $\begin{pmatrix} 1000 & 1000 \\ 100 & 100 \\ 100 & 100 \\ 100 & 100 \end{pmatrix}$ |
| DS3 | 8/2 | $\begin{pmatrix} .125 \\ .125 \\ ... \\ .125 \end{pmatrix}$ | $\begin{pmatrix} 2 & 2 \\ 2 & 2 \\ ... & ... \\ 2 & 2 \end{pmatrix}$ | $\begin{pmatrix} 5 & 5 \\ 10 & 10 \\ ... & ... \\ 640 & 640 \end{pmatrix}$ |
| DS4 | 4/2 | $\begin{pmatrix} .25 \\ .25 \\ .25 \\ .25 \end{pmatrix}$ | $\begin{pmatrix} 1 & 4 \\ 2 & 2 \\ 4 & 1 \\ 2 & 2 \end{pmatrix}$ | $\begin{pmatrix} 100 & 1000 \\ 100 & 1000 \\ 100 & 1000 \\ 100 & 1000 \end{pmatrix}$ |
| DS-CBASP | 3/2 | $\begin{pmatrix} 1/3 \\ 1/3 \\ 1/3 \end{pmatrix}$ | $\begin{pmatrix} 10.9 & 16.2 \\ 9.3 & 19.4 \\ 12.9 & 15.8 \end{pmatrix}$ | $\begin{pmatrix} 99.3 & 79.7 \\ 110.7 & 55.9 \\ 103.5 & 78.6 \end{pmatrix}$ |

Note the similarity of this formulation with (3) in the previous section. As a matter of fact, the optimal solution can be obtained by properly scaling $\sigma_{ij}$, and is given by

$$n_{ij}^* = \frac{v_{ij} \sum_j v_{ij}}{\sum_i (\sum_j v_{ij})^2} N, \qquad (8)$$

where

$$\begin{cases} v_{ij}^2 = \frac{1}{(\mu_{ij^*} - \mu_{ij})^2} \sigma_{ij}^2 & j \neq j^* \\ v_{ij^*}^2 = \sum_{j \neq j^*} \frac{1}{(\mu_{ij^*} - \mu_{ij})^2} \sigma_{ij^*}^2 & j = j^*. \end{cases} \qquad (9)$$

As before, we can use the estimators $\hat{v}_{ij}^2$ to derive an active learning policy, which we call MINMAX-PICS. Note that $j^*$ is unknown as well, and at any time point, it is estimated by the $\text{argmax}_j \hat{\mu}_{ij}$. Thus MINMAXPICS selects the next subpopulation/treatment according to the distribution formed by $\left\{ \frac{\hat{v}_{ij} \sum_j \hat{v}_{ij}}{\sum_i (\sum_j \hat{v}_{ij})^2}; i \in \{1, ..., C\}, j \in \{1, ..., K\} \right\}$.

# 3 Experiments

## 3.1 Minimizing the maximal variance

In this section we compare AREOA's performance with two alternative active learning policies. All start by first sampling each subpopulation and assigning each treatment for a fixed number of times, defined by parameter $B$, and then proceed to a loop of actively selecting the next subpopulation and treatment pair $(i, j)$ until the budget (e.g. $N$) runs out. Full algorithmic details of all three methods are provided in Figure 1. The first alternative, denoted AARandom, recruits subjects from the subpopulation according to the subpopulation fraction in the general population and assigns the treatment uniformly at random. The second alternative, denoted GAFS-MAX, is an active learning policy proposed in [1] for a slightly different loss function: $\max_{1 \leq i \leq C;\ j=1,2} \mathbb{V}[\hat{\mu}_{ij}^n]$. This is similar to the loss in equation (3) however our loss function makes more sense in the context of the clinical trial since the focus is on a relative comparison between mean responses (e.g. the treatment effect), as opposed to the mean responses themselves. In addition to (deterministically) selecting the next (subpopulation, treatment) pair with the highest estimated sample variance, GAFS-MAX also forces a revisiting of (subpopulation, treatment) pairs that haven't been visited for some time.

To illustrate the behaviors of the three active learning policies under different realistic scenarios, we consider the five data sources described in table 1. The response for each subpopulation $i$ under treatment $j$ is modeled using a normal distribution $\mathcal{N}(\mu, \sigma)$. The means and the variances of these normal distributions are detailed in the table. A number of budgets $N$ are considered with the maximum budget set to be 200 in all experiments. For each budget, each algorithm was repeated 100 times with different random initializations of the data sources, so the results reported below are averaged over those 100 runs. Parameter $B$, the initial number of pulls per (subpopulation, treatment) pair is set to 5 for all algorithms. For AREOA, we chose $\epsilon = 0.1$ from $\{0, 0.1, 0.2, 0.3\}$, though the results were consistent for $B = 5$.

For dataset DS1, subpopulation $c_1$ has a large treatment effect variance relative to the other subpopulations. For dataset DS2, the subpopulation distribution is non-uniform in that subpopulation $c_1$ is rare compared to the other subpopulations. For dataset DS3, we consider a scenario where there are 8 subpopulations, with moderate to large variances across subpopulations. For dataset DS4, we consider a scenario where all subpopulations have the same variance in treatment effect, but within each subpopulation there is a large difference in variance between the estimated mean responses to each treatment. For dataset DS-CBASP, the mean and variances are taken from a clinical trial for chronic depression [7]. In this case, patients are stratified based on the severity of their history of alcoholism, and the treatment effect variances of the subpopulations are very similar.

For each dataset, we first plotted the loss of the active learning policies as the budget ($N$) increases (Figure 2). In the figures, the dotted lines at the bottom correspond to the optimal loss of the oracle allocation (i.e. $\frac{\sum_i (\sum_j \sigma_{ij})^2}{N}$, with $N$ varying from 1 to 200).

As shown in Figure 2, for DS1, AREOA utilizes the budget more efficiently than random assignment (AARandom) and GAFS-MAX, particularly for smaller budgets. For DS2, due to the nonuniform distribution of the subpopulation distribution, AARandom performs significantly worse than algorithms that make use of the estimated treatment variances. For DS3, the performance of GAFS-MAX is worse than AARandom, which is suspicious; we discuss this further below. For DS4, we notice that even in cases where there are no significant differences in treatment variance across subpopulations, AREOA still performs quite well; note that as the budget is spent, AREOA approaches the optimal oracle allocation slightly more quickly than AARandom and GAFS-MAX. Recall that in the dataset DS-CBASP, there is little difference in estimated variances across subpopulations and across treatments. In this case, the random assignment policy, AARandom, is perhaps the right choice. We observe that AREOA converges slowly at the beginning but approaches the oracle performance as quickly as AARandom as the budget increases. It is reassuring to see that AEROA performs in a reasonable manner, even for cases which it wasn't specifically designed to handle.

Focusing further on the peculiarity of GAFS-MAX on the dataset DS3, we plotted the allocations and variances of each subpopulation in Figure 3. As is shown in subfigure (a), (b), there are many plateaus for certain high variance subpopulations, which are severely under-explored, due to the fact that resources (allocations) have been devoted to other subpopulations. We think there are a couple of causes behind this behavior. First, when there are many subpopulations, GAFS-MAX will spend time revisiting subpopulations whose treatment variance is well estimated. This can be problematic, especially if the budget is small and there exists some subpopulations that have much larger variances and thus still need more exploration. Another problem with using GAFS-MAX in this setting is the fixed ordering for picking the next rarely visited arm, which may also delay a high variance arm being revisited in the short term. To confirm that the algorithmic behavior of GAFS-MAX is dependent on the ordering of subpopulations, we reversed their ordering in DS3, and we see now in subfigure (c) of Figure 3 that the performance of GAFS-MAX changes significantly.

### 3.2 Minimizing the probability of selecting suboptimal treatments

In this section, we evaluate two variants of the active learning policy MINMAXPICS: MINMAXPICS(SEQ) and MINMAXPICS(GRP) for minimizing the maximal error of selecting a suboptimal treatment. MINMAXPICS(SEQ) is a fully sequential algorithm; it selects the next (subpopulation, treatment) pair proportional to $\{\hat{v}_{ij} \sum_j \hat{v}_{ij}, 1 \leq i \leq C, 1 \leq j \leq K\}$, whereas MINMAXPICS(GRP) selects the next subpopulation proportional to $\{(\sum_j \hat{v}_{ij})^2, 1 \leq i \leq C\}$, and then randomly assigns one patient to each treatment. Note that $\sum_j \hat{v}_{ij} \sum_j \hat{v}_{ij} = (\sum_j \hat{v}_{ij})^2$ for each subpopulation, and we are interested in this variant because it requires fewer interim decision points within a trial, so it may reduce the burden on the trial recruiters. For our algorithms, we also employed a similar epsilon-greedy strategy (with $\epsilon = 0.1$) as that in AREOA. Next, we define a baseline for this problem to be the objective value for (5), calculated by the solutions (8) of the surrogate objective function (7). This baseline provides an estimate of how well MINMAXPICS would perform if it started with the exact prior knowledge and acted greedily according to it.

Table 2: Datasets for the MINMAXPICS comparison

| DS | subpop./ treatments | dist. | means | variances |
|---|---|---|---|---|
| DS21 | 4/3 | $\begin{pmatrix} .25 \\ .25 \\ .25 \\ .25 \end{pmatrix}$ | $\begin{pmatrix} 20 & 10 & 10 \\ 20 & 10 & 10 \\ 20 & 10 & 10 \\ 20 & 10 & 10 \end{pmatrix}$ | $\begin{pmatrix} 50 & 50 & 50 \\ 50 & 50 & 50 \\ 50 & 50 & 50 \\ 50 & 50 & 50 \end{pmatrix}$ |
| DS22 | 4/3 | $\begin{pmatrix} .25 \\ .25 \\ .25 \\ .25 \end{pmatrix}$ | $\begin{pmatrix} 20 & 19 & 15 \\ 20 & 10 & 10 \\ 20 & 10 & 10 \\ 20 & 10 & 10 \end{pmatrix}$ | $\begin{pmatrix} 50 & 50 & 50 \\ 50 & 50 & 50 \\ 50 & 50 & 50 \\ 50 & 50 & 50 \end{pmatrix}$ |
| DS23 | 5/3 | $\begin{pmatrix} .05 \\ .05 \\ .3 \\ .3 \\ .3 \end{pmatrix}$ | $\begin{pmatrix} 20 & 15 & 15 \\ 20 & 15 & 15 \\ ... & ... & ... \\ 20 & 15 & 15 \end{pmatrix}$ | $\begin{pmatrix} 50 & 50 & 50 \\ 50 & 50 & 50 \\ ... & ... & ... \\ 50 & 50 & 50 \end{pmatrix}$ |
| DS24 | 8/3 | $\begin{pmatrix} .125 \\ .125 \\ ... \\ .125 \end{pmatrix}$ | $\begin{pmatrix} 20 & 15 & 15 \\ 20 & 10 & 10 \\ ... & ... & ... \\ 20 & 10 & 10 \end{pmatrix}$ | $\begin{pmatrix} 50 & 50 & 50 \\ 50 & 50 & 50 \\ ... & ... & ... \\ 50 & 50 & 50 \end{pmatrix}$ |
| DS2-CBASP | 3/2 | $\begin{pmatrix} 1/5 \\ 2/5 \\ 2/5 \end{pmatrix}$ | $\begin{pmatrix} 10.9 & 16.2 \\ 9.3 & 19.4 \\ 12.9 & 15.8 \end{pmatrix}$ | $\begin{pmatrix} 99.3 & 79.7 \\ 110.7 & 55.9 \\ 103.5 & 78.6 \end{pmatrix}$ |

As before, we included AARandom in our comparisons. Again, $B$ is set to 5 for all the experiments, so the overall initialization budget varies from dataset to dataset because each dataset may have different number of subpopulations and treatments. The oracle allocation was drawn starting from the budget for initialization to the maximal budget. The details of the datasets used are given in table 2 and the results are presented in Figure 4. The maximal budget is equal to 200 subjects for DS21—DS23. For dataset DS21, in which all subpopulations have the same large treatment effect, all algorithms obtain optimal results even for smaller budgets. Dataset DS22 reflects the setting in which there is a small, but clinically significant, effect between treatments 1 and 2 for one of the subpopulations (here subpopulation 1). Both our active learning policies were able to direct resources quickly to this subpopulation to pick up this effect. For dataset DS23, we assume subpopulations 1 and 2 are rare, so uniform random assignment completely fails to identify the correct treatment. For dataset DS24, we assume there exists many subpopulations, a situation that could happen with complex stratification based on multiple biomarkers. Also for better displaying the trend, the maximum budget is increased to 250 for this dataset. One of the subpopulations has slightly smaller treatment effect, again the uniform random assignment failed to identify the correct treatment because the resources are spread too thin. As before, the DS2-CBASP example is modeled after a clinical dataset with three subpopulations [7]. The budget is set to 700 mimicking 680, the number of patients in the original trial. we also assume that subpopulation 1 is slightly rare. We observe advantage of our methods when the budget is limited. For example, it takes about 255 subjects to reach 20% error rate, while it takes almost 500 subjects for AArandom. We also see that oracle allocations indicate that there is a lot of room for improvement with prior knowledge. Overall, MINMAXPICS-SEQ and MINMAXPICS-GRP performed about the same.

# 4 Related Work

This problem addressed in this paper is related to the famous "multi-armed bandit problem" by Robbins [11]. For example, in clinical trials, our formulation bears formal similarity with some of response adaptive trials [14], popular in cancer research, which also divide patients into groups. These trials aim to place more subjects on what appears to be the better treatment in each group. A key difference between our work and the conventional multi-armed bandit problem is that in the latter, one tries to maximize the cumulative rewards over all pulls, whereas with this work, one simply wants to maximize a measure of the quality of the resulting ITRs once the budget is used up. Problems with similar interests in only the "end results" have been studied under the names of "budgeted" multi-armed bandit problems [9, 4]. Budgeted learning is related to this problem in several ways. First, both have a "hard" budget constraint. Second, the performance depends on the quality of the *final* decision once the budget is exhausted. For our first objective in Section 2.1, the closest work to ours is that of Antos et al.[1] with a similar goal of reducing the variances of the estimated mean values.

Our work is also related to the topic of active learning, which has a long history in machine learning [3, 12] as well as in statistics; in the latter this is generally referred to as optimal experimental design [5]. The main objective of this line of research is to reduce the variance of prediction over parameter estimates, while controlling the bias of the prediction at the same time. Our problem shares some similarity with this literature in that our first formulation is to minimize the maximal variance of the estimated treatment effects.

Our second optimization criterion (Section 2.2), is related to the topic of ranking and selection (R&S) in operations research [8]. In particular, our allocation scheme is related to a popular method called OCBA [2] for R&S problems in the setting in which there is only one subpopulation. Essentially, ranking and selection is a form of budgeted learning with 0-1 loss functions, though the focus there is on the minimally required sample size and the statistical soundness of the online procedures.

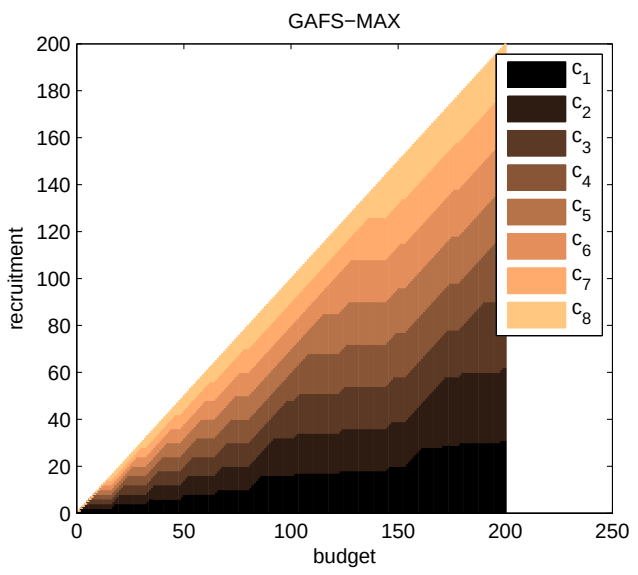

(a) gafs-max subpopulation allocations

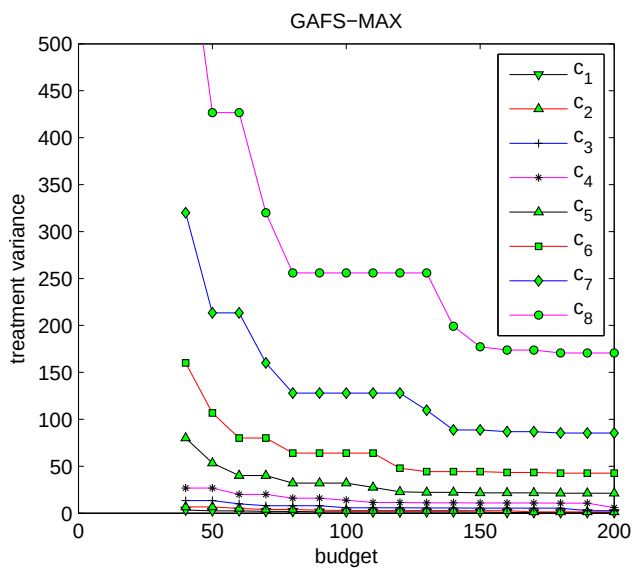

(b) gafs-max subpopulation variances

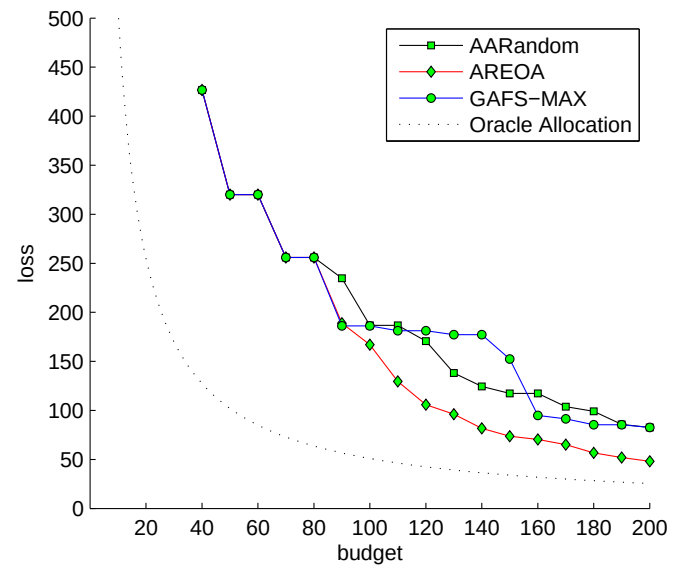

(c) DS3 with subpopulation ordering reversed

Figure 3: Algorithm GAFS-MAX on DS3, subfigure (a), (b) are the subpopulation allocations and variances, subfigure (c) uses the same dataset as DS3, with ordering of the subpopulations reversed

## 5 Summary and Future Work

We presented an active learning approach for acquiring data with the goal of learning individualized treatment rules. The first optimization criterion aims to bound the worse case uncertainty of the ITR for any subpopulation, while the second optimization criterion aims to bound the worse case error of picking a suboptimal treatment for any subpopulation. For both optimization criteria discussed, we demonstrated the potential of active learning policies for cost-saving, in comparison with a completely randomized exploration policy. Currently, the total budget is expressed as a parameter ($N$). It would be useful to provide a way to estimate the required total budget so as to ensure that the budget is not too small to provide a high quality ITR. One potential issue with selecting a single best treatment (Section 2.2) is that when there is no clear winner among treatments for a subpopulation, our approach would still devote excessive resources to it. One way to avoid this is to incorporate a stopping rule, causing the approach to "give up" on this subpopulation. Another way is to modify the objectives to select any near optimal treatment not far away from the best treatment. Pertaining to this idea, yet another meaningful objective is to identify as many good treatments for all subpopulations under the budget constraint. Finally, "dynamic treatment regimes" or "adaptive treatment strategies" [10, 6] naturally generalize the idea of ITRs to multiple stages by constructing a sequence of decision rules, one for each disease stage. The extension of the ideas presented in this paper to such time-varying settings is also an interesting avenue for future work.

## Acknowledgements

We acknowledge support from the National Institutes of Health (NIH) grants R01 MH080015 and P50 DA10075.

## References

[1] A. Antos, V. Grover, and C. Szepesvári. Active learning in multi-armed bandits. In *Proceedings of the 19th international conference on Algorithmic Learning Theory*, pages 287–302, Budapest, Hungary, 2008. Springer-Verlag.

[2] C. H. Chen, J. Lin, E. Yücesan, and S. E. Chick. Simulation budget allocation for further enhancing the efficiency of ordinal optimization. *Discrete Event Dynamic Systems*, 10(3):251–270, 2000.

[3] D. Cohn, L. Atlas, and R. Ladner. Improving generalization with active learning. *Machine Learning*, 15(2):201–221, 1994.

[4] K. Deng, C. Bourke, S. Scott, J. Sunderman, and Y. Zheng. Bandit-based algorithms for budgeted learning. In *ICDM 2007*, page 463–468, 2008.

[5] V. Fedorov. Optimal experimental design. *Wiley Interdisciplinary Reviews: Computational Statistics*, 2010.

[6] A. Guez, R. D. Vincent, M. Avoli, and J. Pineau. Adaptive treatment of epilepsy via batch-mode reinforcement learning. In *Proceedings of the Twentieth Innovative Applications of Artificial Intelligence Conference*, page 1671–1678, 2008.

[7] M. B. Keller, J. P. McCullough, D. N. Klein, B. Arnow, D. L. Dunner, A. J. Gelenberg, J. C. Markowitz, C. B. Nemeroff, J. M. Russell, and M. E. Thase. A comparison of nefazodone, the cognitive behavioral-analysis system of psychotherapy, and their combination for the treatment of chronic depression. *New England Journal of Medicine*, 342(20):1462, 2000.

[8] S. H. Kim and B. L. Nelson. Recent advances in ranking and selection. In *Proceedings of the 39th conference on Winter simulation: 40 years! The best is yet to come*, page 162–172, 2007.

[9] O. Madani, D. J. Lizotte, and R. Greiner. The budgeted multi-armed bandit problem. *Colt 2004*, 3120:643—645, 2004.

[10] S. A. Murphy, K. G. Lynch, D. Oslin, J. R. McKay, and T. TenHave. Developing adaptive treatment strategies in substance abuse research. *Drug and alcohol dependence*, 88:S24–S30, 2007.

[11] H. Robbins. Some aspects of the sequential design of experiments. *Bulletin of the American Mathematical Society*, 58(5):527–535, 1952.

[12] B. Settles. Active learning literature survey. Computer Sciences Technical Report 1648, University of Wisconsin–Madison, 2009.

[13] M. R. Trusheim, E. R. Berndt, and F. L. Douglas. Stratified medicine: strategic and economic implications of combining drugs and clinical biomarkers. *Nat Rev Drug Discov*, 6(4):287–293, Apr. 2007.

[14] X. Zhou, S. Liu, E. S. Kim, R. S. Herbst, and J. J. Lee. Bayesian adaptive design for targeted therapy development in lung cancer–a step toward personalized medicine. *Clinical Trials (London, England)*, 5(3):181–193, 2008. PMID: 18559407.

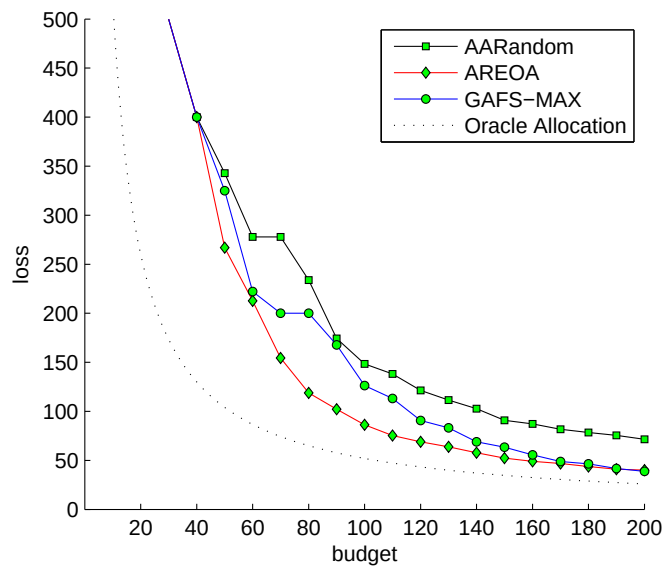

(a) DS1

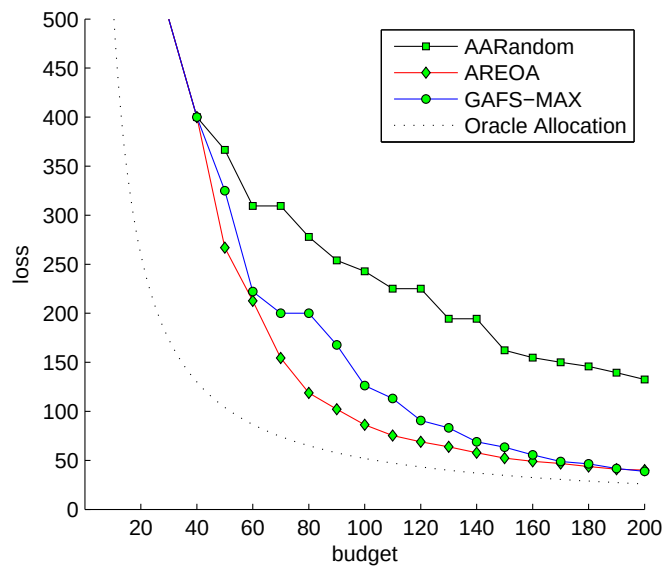

(b) DS2

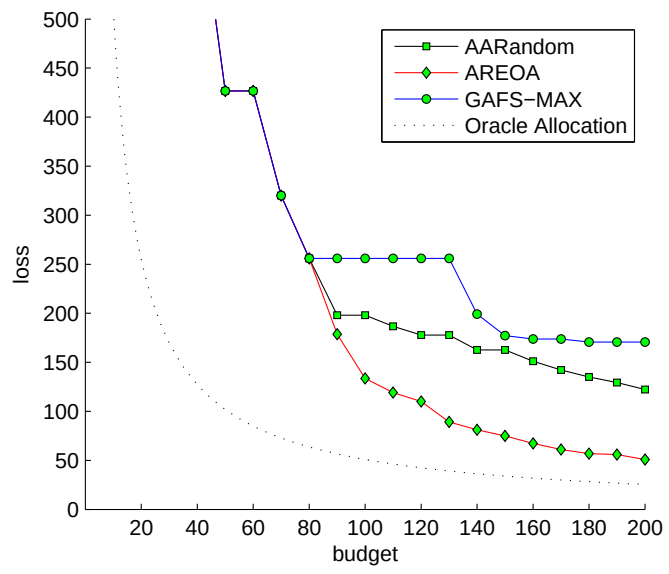

(c) DS3

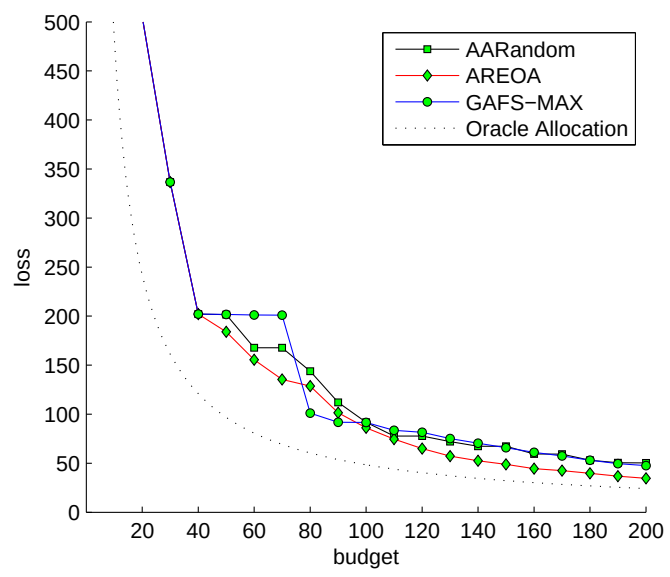

(d) DS4

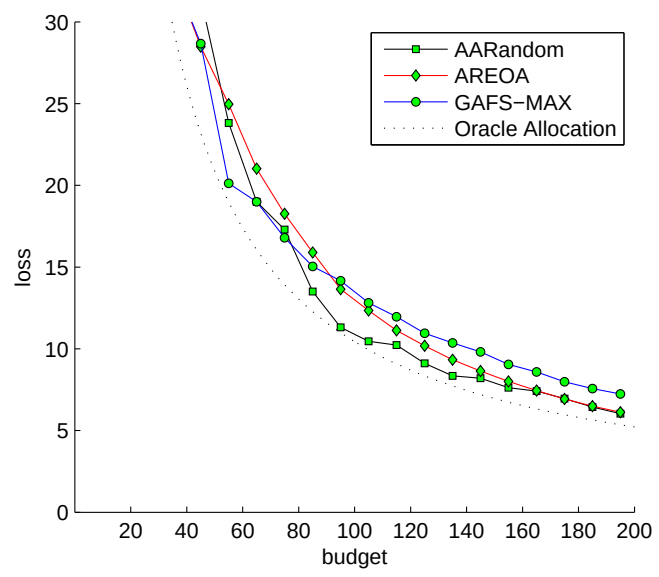

(e) DS-CBASP

Figure 2: Simulation results for the objective of minimizing overall variance

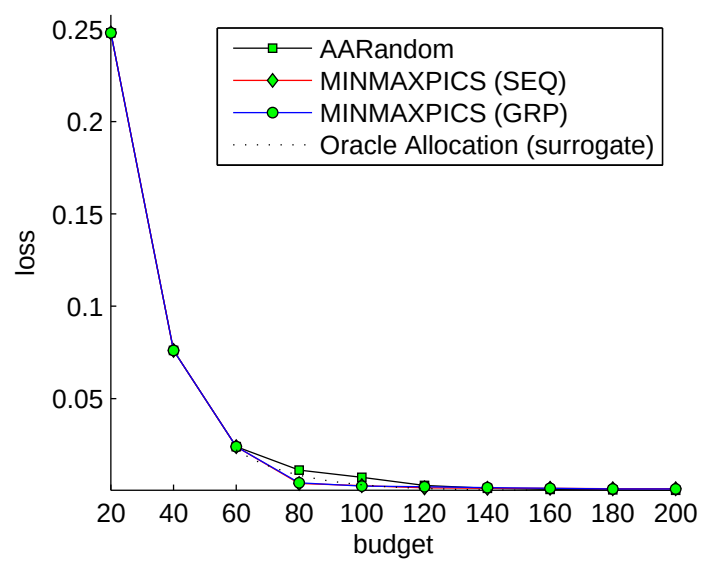

(a) DS21

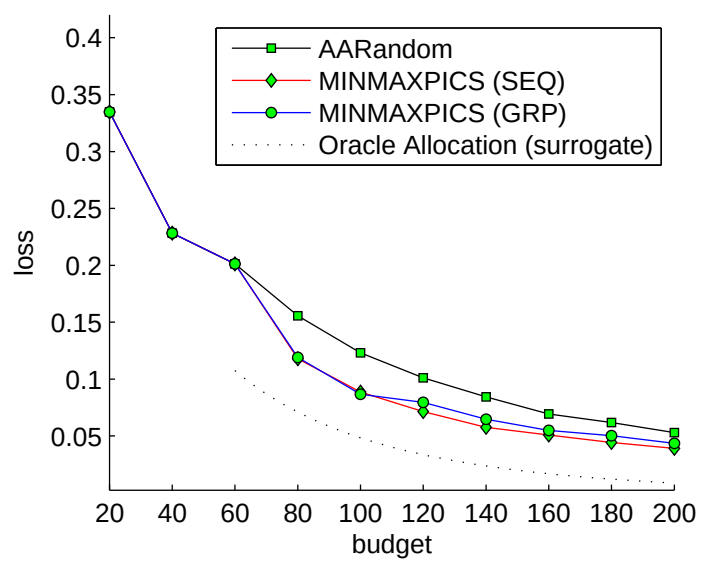

(b) DS22

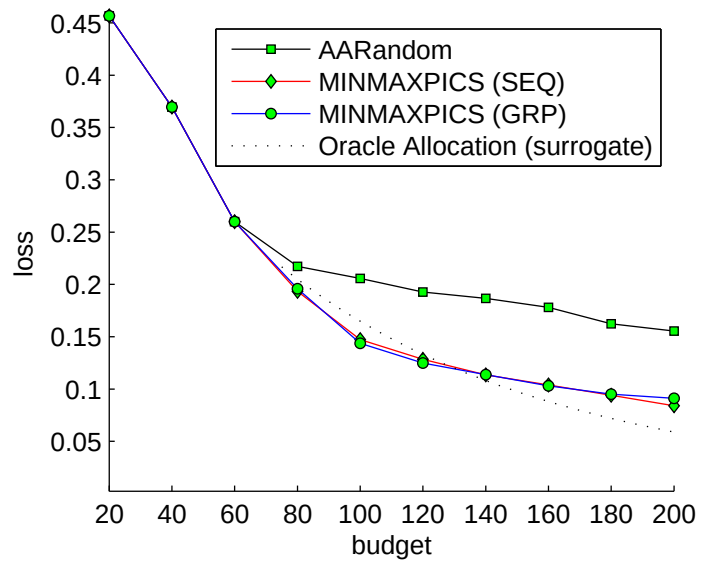

(c) DS23

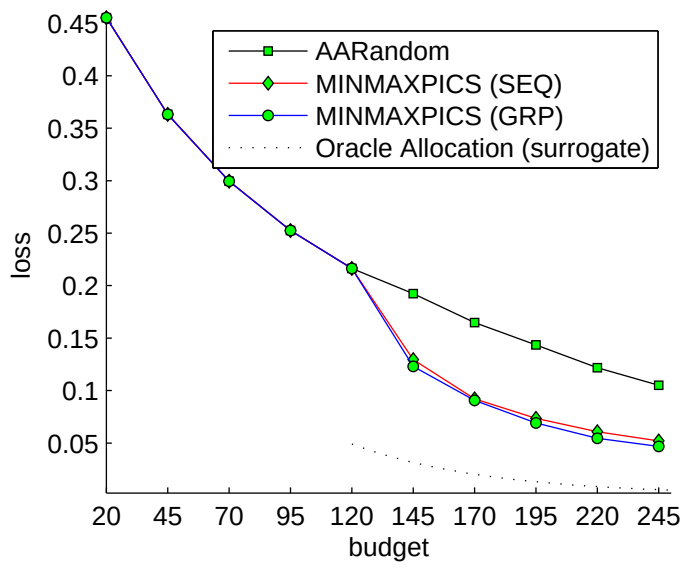

(d) DS24

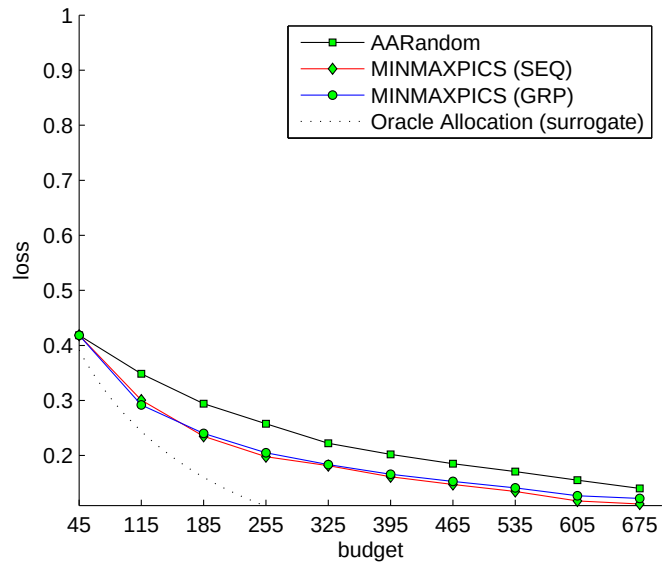

(e) CBASP-2

Figure 4: Simulation results for the objective of minimizing overall error of selecting suboptimal treatment